\documentclass{article}
\usepackage{spconf,amsmath,amssymb,graphicx,url}
\usepackage{tikz}
\usepackage{comment}
\usepackage[
    backend=biber,
    style=ieee,
    maxbibnames=3,
    maxcitenames=3,
    doi=false,isbn=false,url=false,eprint=false
]{biblatex} 

\newcommand{\Fig}[1]{Figure~\ref{fig:#1}} 
\newcommand{\Sec}[1]{Section~\ref{sec:#1}} 

\newcommand{\drawfig}[4]{ 
    \begin{figure}[#1]
    \centering 
    \includegraphics[width=#2]{#3} 
    \caption{#4}
    \label{fig:#3}
    \end{figure}
}
\newcommand{\drawfigfull}[4]{ 
    \begin{figure*}[#1]
    \centering 
    \vspace{0mm}
    \includegraphics[width=#2]{#3} 
    \caption{#4}
    \label{fig:#3}
    \end{figure*}
}

\DeclareSourcemap{
	\maps[datatype=bibtex, overwrite=true]{
		\map{
			\step[fieldsource=booktitle,
			match=\regexp{.*Interspeech.*},
			replace={Proc. Interspeech}]
			\step[fieldsource=journal,
			match=\regexp{.*INTERSPEECH.*},
			replace={Proc. Interspeech}]
			\step[fieldsource=booktitle,
			match=\regexp{.*icassp_inpress.*},
			replace={ICASSP (in press)}]
			\step[fieldsource=booktitle,
			match=\regexp{.*International.*Conference.*on.*Acoustics,.*Speech.*and.*Signal.*Processing.*},
			replace={Proc. ICASSP}]
			\step[fieldsource=booktitle,
			match=\regexp{.*International.*Conference.*on.*Learning.*Representations.*},
			replace={ICLR}]
			\step[fieldsource=booktitle,
			match=\regexp{.*International.*Conference.*on.*Machine.*Learning.*},
			replace={ICML}]
			\step[fieldsource=booktitle,
			match=\regexp{.*Automatic.*Speech.*Recognition.*and.*Understanding.*},
			replace={Proc. ASRU}]
			\step[fieldsource=booktitle,
			match=\regexp{.*Spoken.*Language.*Technology.*},
			replace={Proc. SLT}]
			\step[fieldsource=booktitle,
			match=\regexp{.*Speech.*Synthesis.*Workshop.*},
			replace={Proc. SSW}]
			\step[fieldsource=booktitle,
			match=\regexp{.*workshop.*on.*speech.*synthesis.*},
			replace={Proc. SSW}]
			\step[fieldsource=booktitle,
			match=\regexp{.*Advances.*in.*neural.*information.*processing.*},
			replace={Proc. NIPS}]
			\step[fieldsource=booktitle,
			match=\regexp{.*Advances.*in.*Neural.*Information.*Processing.*},
			replace={Proc. NIPS}]
			\step[fieldsource=booktitle,
			match=\regexp{.*Workshop.*on.* Applications.* of.* Signal.*Processing.*to.*Audio.*and.*Acoustics.*},
			replace={Proc. WASPAA}]
			\step[fieldsource=booktitle,
			match=\regexp{.*International.*Conference.*on.*Language.*Resources.*and.*Evaluation.*},
			replace={Proc. LREC}]
			\step[fieldsource=journal,
			match=\regexp{.*Spontaneous.*Speech.*Processing.*and.*Recognition},
			replace={Proc. SSPR}]
			\step[fieldsource=publisher,
			match=\regexp{.+},
			replace={{}}]
			\step[fieldsource=month,
			match=\regexp{.+},
			replace={{}}]
			\step[fieldsource=location,
			match=\regexp{.+},
			replace={{}}]
			\step[fieldsource=address,
			match=\regexp{.+},
			replace={{}}]
			\step[fieldsource=organization,
			match=\regexp{.+},
			replace={{}}]
			\step[fieldsource=doi,
			match=\regexp{.+},
			replace={{}}]
			\step[fieldsource=url,
			match=\regexp{.+},
			replace={{}}]
			\step[fieldsource=editor,
			match=\regexp{.+},
			replace={{}}]
		}
	}
} 

\title{Do learned speech symbols follow Zipf's law?}
\name{Shinnosuke Takamichi, Hiroki Maeda, Joonyong Park, Daisuke Saito, and Hiroshi Saruwatari\thanks{This work is supported by JSPS KAKENHI 21H04900 (practical experiment) and Moonshot R\&D Grant Number JPMJPS2011 (algorithm development).}}
\address{The University of Tokyo, Japan.}

\begin{document}
\ninept
\maketitle
\setlength{\tabcolsep}{1mm} 

\begin{abstract}
    In this study, we investigate whether speech symbols, learned through deep learning, follow Zipf's law, akin to natural language symbols. Zipf's law is an empirical law that delineates the frequency distribution of words, forming fundamentals for statistical analysis in natural language processing. Natural language symbols, which are invented by humans to symbolize speech content, are recognized to comply with this law. On the other hand, recent breakthroughs in spoken language processing have given rise to the development of learned speech symbols; these are data-driven symbolizations of speech content. Our objective is to ascertain whether these data-driven speech symbols follow Zipf's law, as the same as natural language symbols. Through our investigation, we aim to forge new ways for the statistical analysis of spoken language processing. 
\end{abstract} \vspace{-1mm}


\section{Introduction} \label{sec:introduction}
    Zipf's law, a well-known empirical principle, delineates the frequency of occurrence of elements within a dataset~\cite{zipf49human}. Specifically, when the occurrence frequency of an element ranks as the $k$-th highest within a dataset, it equates to $1/k$ of the frequency of the most occurring element. This law is observed to be applicable across various data domains, with natural language symbols (e.g., words), follow this pattern~\cite{steven14zipflawnlp}. To illustrate, the third most frequent word in an English document, ``and'', appears approximately one-third as often as the most frequent word, ``the''\footnote{\url{https://www.cs.cmu.edu/~cburch/words/top.html}}. In the practice of analyzing the occurrence frequency of natural language symbols within a text corpus, one can denote the frequency rank as $r$ and its respective occurrence frequency as $f_r$. Consequently, the following relationship is given by Zipf's law:
        \begin{align}
            f_r = a r^{-\eta},
        \end{align}
    where, $a$ and $\eta$ are model parameters. The text corpus follows Zipf's law when $\eta \approx 1$, and follows a power law otherwise. In essence, Zipf's law is a specific type of power law. When following a power law, the log-log plot of rank against frequency appears linear.

    By examining the adherence to or deviation from Zipf's law, one can analyze the distinct characteristics of a text corpus. This analysis finds notable applications in natural language processing, as illustrated by the following examples~\cite{tanaka21statisticaluniversals}:
    \begin{itemize} \leftskip -8mm 
        \item[] \textbf{Infants' language acquisition}: The vocabulary that 2- to 4-year-olds acquire tends to gravitate towards high-frequency words, a phenomenon indicated by a rank-frequency distribution that is convex, rather than linear~\cite{elena09twoyearoldproduction}.
        \item[] \textbf{Writing system variations across languages}: Writing systems significantly differ from one language to another, ranging from sound-based phonograms to semantic-based logograms. Tanaka observed that as notation shifts from phonogram to logogram, the rank-frequency distribution transitions to follow a power law; phonographic languages display a convex, while logographic languages demonstrate a linear on a plot~\cite{tanaka21statisticaluniversals}.
        \item[] \textbf{Quantification of communication effort}: Zipf's law is also recognized as the \textit{principle of least effort}~\cite{linders23zipf2lawrevisited}. This principle stipulates that the utilization of frequently used words minimizes speaking and listening efforts during human communication. Viewing communication through this perspective allows for the potential measurement of the naturalness exhibited in communication, including emergent communication~\cite{kottur-etal-2017-natural} and machine-generated communication~\cite{serhii17nips_zipf}.
    \end{itemize} \vspace{-1mm}
    
    Meanwhile, recent advancements in deep learning, such as self-supervised learning, have facilitated the discovery of discrete symbol representations~\cite{lakhotia-etal-2021-generative,park23interspeech_gslm-investigation,borsos22audiolm}. These representations are learned in a data-driven manner from speech. In contrast to traditional signal-processing-based representations, such as mel-spectrograms, the symbol sequences learned through these methods captures a wealth of phonetic and semantic contents. While natural language symbols were \textit{crafted by humans} to encode the contents of speech, the learned speech symbols can be seen as their data-driven counterparts, that means, symbols crafted in a data-driven manner.

    With this context in mind, we are led to a research question:
        \begin{center}
            ``Does Zipf's law, which holds true for natural language symbols, also apply to the learned speech symbols?''
        \end{center}
    Verifying this hypothesis could potentially unveil the capability to extend the statistical analysis techniques, traditionally employed in natural language analysis, to the spoken language analysis. Moreover, since this methodology bypasses the necessity for transcriptions, it might forge a path towards textless analysis applicable to a diverse range of sounds, encompassing non-verbal vocalizations and non-speech sounds (further details are elaborated in \Sec{conclusion}).

    In this study, we conduct experiments to address the posed question, utilizing the generative spoken language model (GSLM)~\cite{lakhotia-etal-2021-generative}, a variant of speech symbol representation methodology. Initially, we perform a foundational experiment to ascertain whether the learned speech symbols follow Zipf's law. Leveraging a speech corpus comprising paired text and speech, we analyze symbol rank-frequency distributions by correlating them with the accompanying text. Subsequently, we explore whether our textless analysis identifies deviation in non-textual contents of speech. Specifically, we aim to capture variations in language fluency between native and non-native speech utterances. Our efforts endeavor to pave a way for comprehensive speech analysis.


\section{Generative spoken language model (GSLM)} \label{sec:related-work}
    \drawfig{t}{0.45\linewidth}{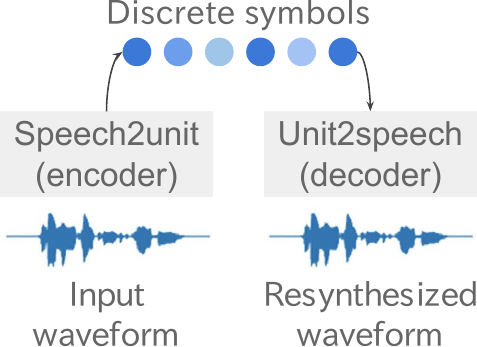}
    {Generative spoken language model.}

    \Fig{fig/gslm.pdf} illustrates the GSLM~\cite{lakhotia-etal-2021-generative}, an analysis-synthesis system operating through discrete speech symbols. This system is structured into three modules: speech2unit, unit language model, and unit2speech. However, this study focuses solely on the utilization of the speech2unit. This specific module integrates a pre-trained self-supervised learning (SSL) model, featuring technologies such as contrastive predictive coding~\cite{abs-1807-03748}, wav2vec2.0~\cite{NEURIPS2020_92d1e1eb}, HuBERT~\cite{9585401}, in conjunction with a $k$-means clustering model. The SSL model remains fixed, while the $k$-means clustering model is trained using a speech corpus. Within this setup, the speech2unit module transforms a speech waveform into a sequence of discrete symbols.
    
    Symbols are generated at intervals of $20$~ms, this interval generally shorter than the phoneme duration. Consequently, the speech2unit frequently predicts identical symbols. To prevent this redundancy, we have opted to consolidate consecutive identical symbol sequences into a single representation. For instance, if the speech2unit outputs a sequence like $[3, 3, 3, 50, 200, 200]$ (where the numbers indicate symbol indices), we simplify this to $[3, 50, 200]$. It's important to note that the speech2unit, particularly the $k$-means clustering model, is highly sensitive to language of input speech. Hence, it's imperative that the model and the encoded speech align in terms of language.

\section{Methodology} \label{sec:model_construction}
    We delineate two methodologies employing GSLM speech symbols to explore Zipf's law.
    
    \textbf{Examining the applicability of Zipf's law to speech symbols.}
    This methodology is conducted utilizing text-speech pairs. In the case of text, we employ natural language symbols, represented either as words or as character $n$-grams. Words are extracted through morphological analysis to discern the original form of the words, while character $n$-grams involve sequences of $n$ consecutive characters. For speech, we obtain speech symbols as outlined in \Sec{related-work}, calculate speech symbol $n$-grams, that is, sequences of $n$ consecutive speech symbols. We then calculate the ratio of the lengths of natural language symbol sequences to speech symbol sequences to determine the value of $n$. To describe this intuitively, $n$ signifies the average number of speech symbols corresponding to a single natural language symbol. We then assess whether Zipf's law holds true for speech symbol $n$-grams within text adhering to the law.
    
    \textbf{Identifying non-textual deviations from Zipf's law.}
    As explained in \Sec{introduction}, identifying deviations from this law can be a method to pinpoint non-standard word usage. In this methodology, we explore to detect non-standard speech patterns based on rank-frequency distribution. By comparing distributions of standard and non-standard speech, we aim to identify deviations in non-standard speech from the standard speech. 

\section{Experimental evaluation} \label{sec:experiment}
    \subsection{Experimental condition}
        We utilized HuBERT~\cite{9585401}, which was trained on LibriSpeech~\cite{panayotov15librispeech}, as the SSL model in the GSLM speech2unit. The language-specific $k$-means clustering models were trained using JSUT/JVS~\cite{takamichi2020jsut}, J-KAC~\cite{nakata21_ssw}, and J-MAC~\cite{takamichi22_interspeech} for Japanese, and LibriSpeech for English. The number of classes was set to $200$ for both languages. The models, implemented using the \texttt{fairseq} toolkit~\cite{ott-etal-2019-fairseq}, are publicly available\footnote{\url{https://huggingface.co/nonmetal/gslm-japanese} (Japanese)\cite{park23asjs_gslm}}\footnote{\url{https://github.com/facebookresearch/fairseq/tree/main/examples/textless_nlp/gslm} (English)}. The speech sampling frequency was set at $16$ kHz, and speech symbols were extracted every $20$ ms. Given that high-frequency and low-frequency items often deviate from Zipf's law\cite{tanaka21statisticaluniversals}, we estimated model parameters $a$ and $\eta$ based only on the top $0.1$~\% to $10$~\% of the frequencies. These model parameters were determined using the least square method. For comparative analysis, we fixed $\eta$ at $1.0$ while estimating only $a$, indicating a strict adherence to Zipf's law in the rank-frequency distribution. To reduce the data size of figures, we thinned the data to be plotted.

        To verify Zipf's law in \Sec{experiment-zipf}, we utilized approximately $7,600$ Japanese utterances from JSUT~\cite{takamichi2020jsut} and $13,000$ English utterances from LJSpeech~\cite{panayotov15librispeech}. For word tokenization, MeCab\footnote{\url{https://taku910.github.io/mecab/}} and NLTK\footnote{\url{https://www.nltk.org/}} served as morphological analyzers. In the character $n$-gram analysis, we used a set of characters encompassing Chinese/Japanese characters and marks for Japanese, and lowercase alphabets, symbols (e.g., ":", "?"), and whitespace for English. The average number of characters per word was $1.6$ (ja) and $5.1$ (en), while the average number of speech symbols per character stood at $5.7$ (ja) and $1.9$ (en). Moreover, the average number of speech symbols per word was $8.9$ (ja) and $9.0$ (en). The value of $n$ in both character $n$-gram and speech symbol $n$-gram was set to the ceiling of these values, for instance, $\lceil 1.6 \rceil = 2$ for Japanese characters.
        
        For identifying non-textual deviation in \Sec{experiment-non-native}, we employed native and Japanese-accented English utterances from UME-ERJ\footnote{\url{https://research.nii.ac.jp/src/en/UME-ERJ.html}}. The corpus contained about $20$ native and $200$ non-native speakers, each reading approximately 300--500 English sentences. These sentences varied between speakers but maintained a balanced phoneme distribution. Non-native speakers were assigned a language fluency score on a five-point scale. Based on these scores, we categorized non-native speakers into three groups: low- ($\mathrm{score} < 3.0$), mid- ($3.0 \leq \mathrm{score} < 3.5$), and high-level ($\mathrm{score} \geq 3.5$). These groups consisted of $74$, $66$, and $46$ speakers, respectively. We aggregated speech symbols separately for each non-native speaker group and the native speakers. To equalize data size across groups, we randomly selected $10,000$ utterances per group. For symbol encoding, we used the English GSLM speech2unit. 

    \subsection{Verifying Zipf's law of speech symbols} \label{sec:experiment-zipf}
        We verify the law via three steps: word, character $n$-gram, and speech symbol $n$-gram.
    
        \subsubsection{Word}
            First, we verify that the words in the corpora we used adhere to Zipf's law. \Fig{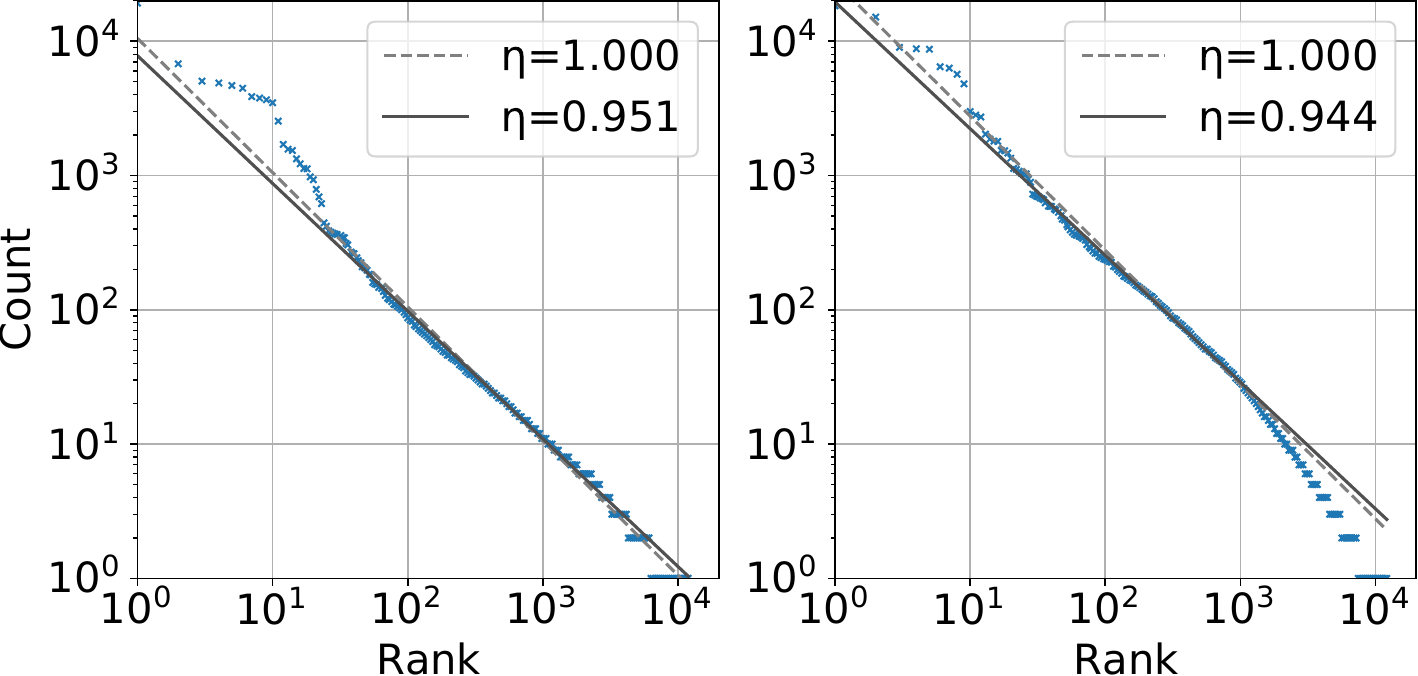} illustrates the rank-frequency distributions in both Japanese and English. These distributions appear to be linear and the value of $\eta$ is close to $1.0$. Therefore, we can say that the words follow Zipf's law, and that the corpora used have the universal statistics in terms of the word distribution. A minor observation, as highlighted in previous studies~\cite{tanaka21statisticaluniversals}, is that the high-frequency ($\mathrm{rank} < 20$) and low-frequency ($10^3 < \mathrm{rank}$) items deviate from the regression line.
    
            \drawfig{t}{0.98\linewidth}{fig/zipf_word.pdf}
            {Word rank-frequency distributions. left: Japanese, right: English.}

        \subsubsection{Character $n$-gram}
            Next, we explore another natural language symbol: the character $n$-gram. \Fig{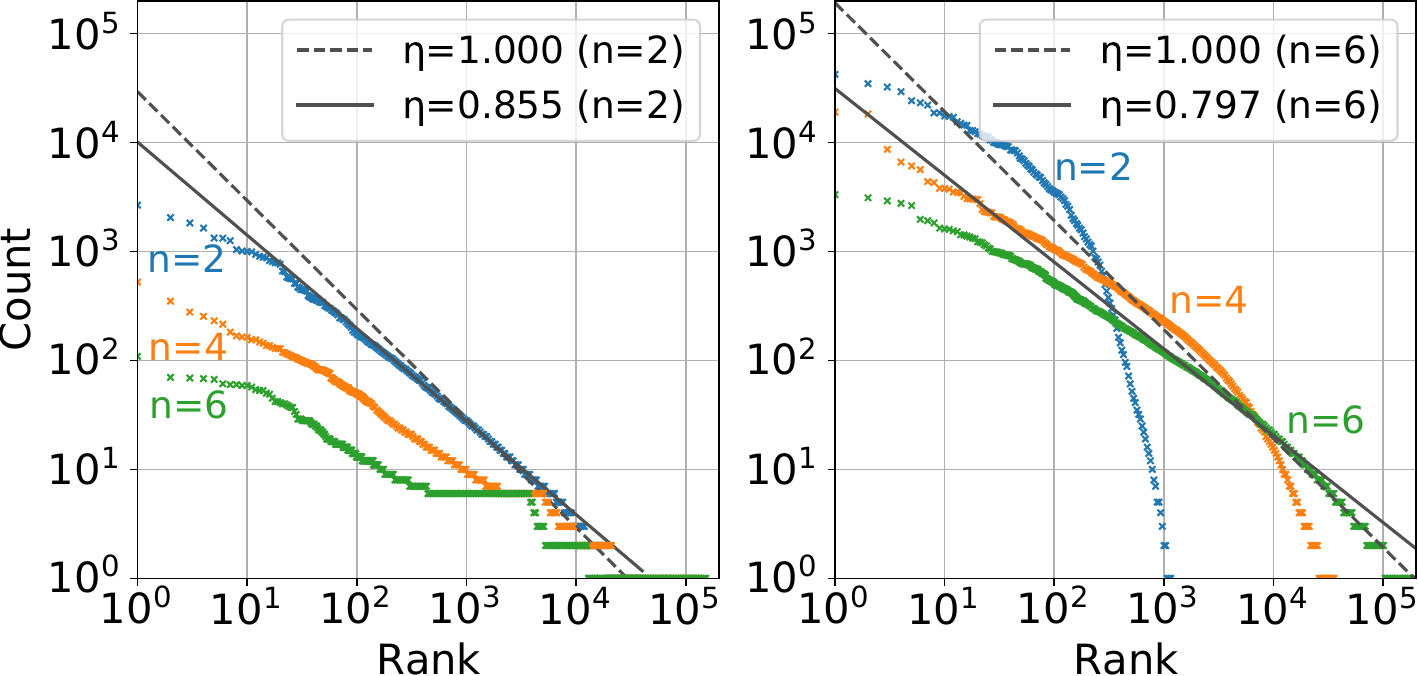} shows these distributions. When analyzed for the same value of $n$, differences between languages become evident. In Japanese, the distribution attains linear at $n = 2$, whereas in English, it displays convex. The English distribution gradually transitions to a linear shape as $n$ increases, and at $n = 6$, it remains convex but closely resembles a linear trend. At $n = 2$ for Japanese and $n = 6$ for English that correspond to a word, the distributions of character $n$-grams more closely follow a power law, rather than Zipf's law.
        
            \drawfig{t}{0.98\linewidth}{fig/zipf_char.pdf}
            {Character $n$-gram rank-frequency distributions. left: Japanese, right: English. $n=2$ in Japanese and $n=6$ in English correspond a word.}
        
        \subsubsection{Speech symbol $n$-gram}
            Finally, we explore whether speech symbols follow Zipf's law by comparing the results with those of words and character $n$-grams. \Fig{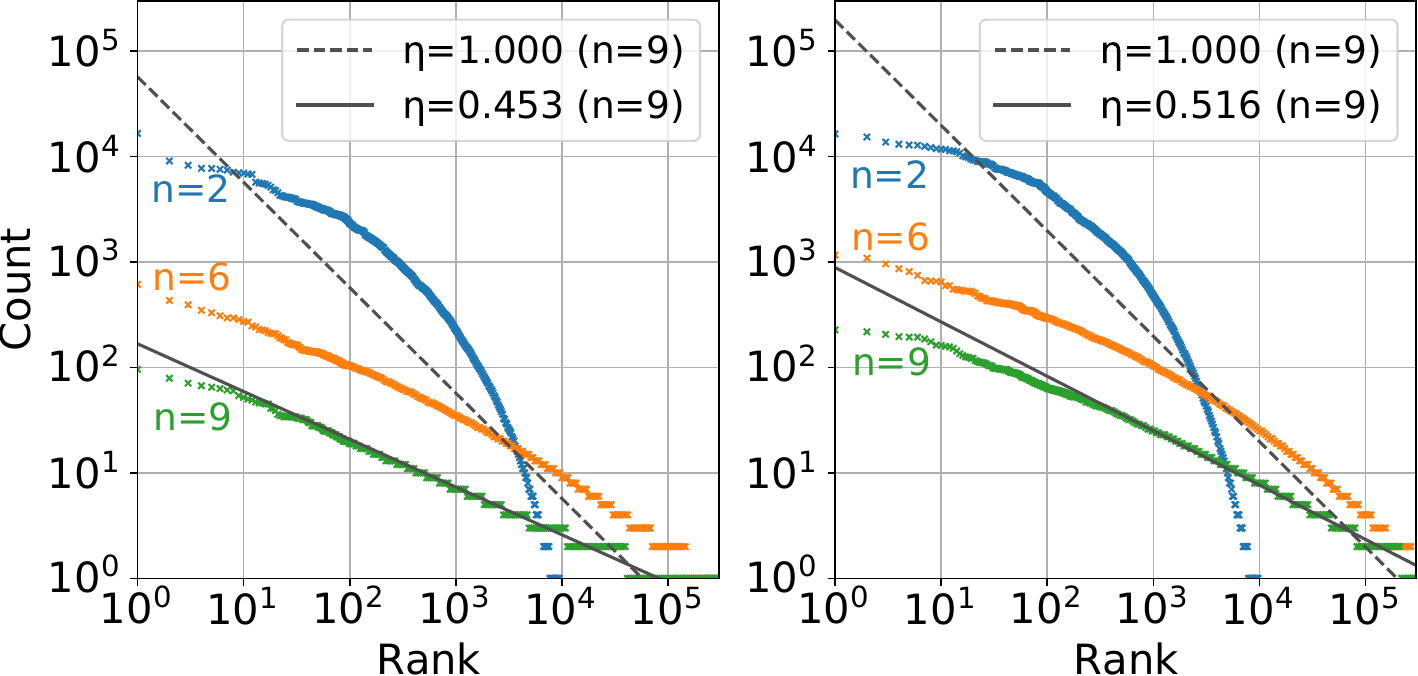} shows these distributions.
            
            \textbf{Difference between languages.}
            When analyzed for the same value of $n$, we find that the distributions are almost identical. These results suggest that the distributions of consecutive speech symbols, which indicate the frequency of speech segment use, are language-independent, at least between Japanese and English. The distributions for English are slightly shifted upwards compared to those for Japanese. This discrepancy is due to the difference in data size; the English corpus is twice as large as the Japanese corpus.
            
            \textbf{Comparison in $n$ corresponding to a character.}
            $n=6$ for Japanese and $n=2$ for English corresponds to a character. We observe that distributions where $n$ corresponds to a character maintain shapes similar to those of character 1-gram. Specifically, the distribution for $n=6$ in Japanese is nearing linearity (but not $\eta = 1.0$), while the distribution for $n=2$ in English is convex (Although not illustrated in \Fig{fig/zipf_char.pdf}, the distributions of character 1-grams resemble those of the 2-gram distributions: they are linear for Japanese and convex for English.). This difference can be traced back to differences in writing systems, as discussed in \Sec{introduction}. Namely, Japanese and English are more akin to logographic and phonographic languages, respectively. The closer a symbol is to representing semantic information rather than phonetic one, the more linear the distribution becomes. Our findings suggest that this trend holds true for speech symbols and that the speech symbol $n$-gram might statistically reflect the sound or meaning of a character. The ability to attain character statistics without relying on characters also facilitates analysis based on Zipf's law (and power law) without involving characters.
            
            \textbf{Comparison in $n$ corresponding to a word.}
            $n=9$ signifies for both Japanese and English. As $n$ escalates, the distributions gravitate towards linear (not $\eta = 1.0$). We theorize that this phenomenon occurs because, as noted previously, $n$-grams tend to represent semantic information as $n$ increases. $\eta$ not being $1.0$, but the distribution at $9$-grams is linear, mirroring the word distribution observed in \Fig{fig/zipf_word.pdf}. This suggests that our methodology holds promise for statistically analyzing features associated with words without utilizing words themselves.    
        
            \drawfig{t}{0.98\linewidth}{fig/zipf_symbol.pdf}
            {Speech symbol $n$-gram rank-frequency distributions. left: Japanese, right: English. $n=6$ in Japanese and $n=2$ in English correspond to a character. $n=9$ corresponds a word.}

    \subsection{Identifying non-textual deviation} \label{sec:experiment-non-native}
        \drawfigfull{t}{0.98\linewidth}{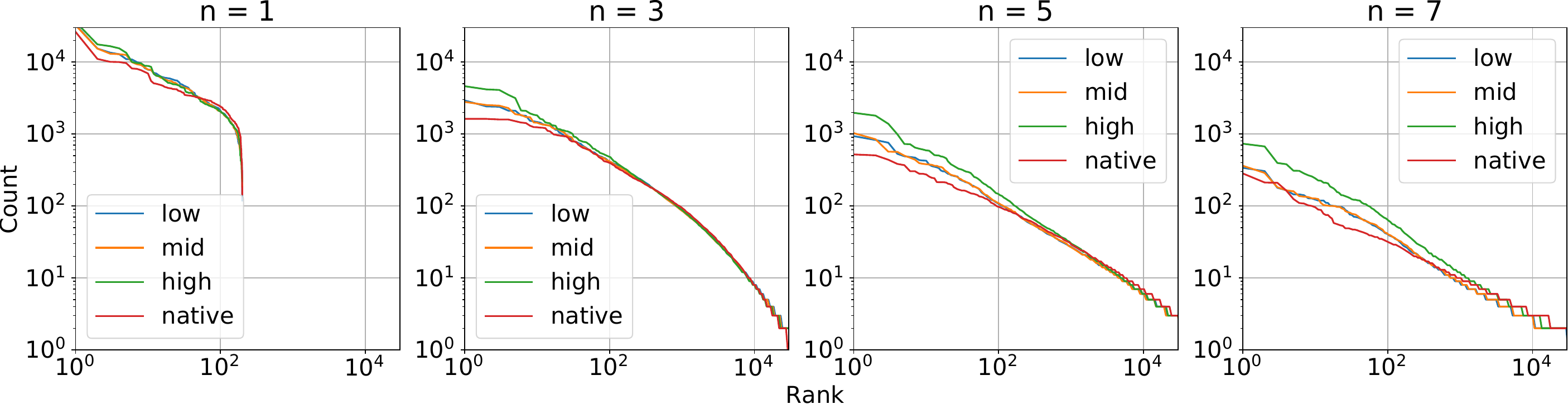}
        {Speech symbol $n$-gram frequency of native and non-native speech. ``low,'' ``mid,'' and ``high'' indicate English fluency level of non-native speakers. For clear illustration, we choose line plots rather than scatter plots.}

        \Fig{fig/non-native.pdf} displays the speech symbol $n$-gram ($n={1,3,5,7}$) frequencies in both native and non-native speakers. We delve into the distinctions between native and non-native speakers based on the data presented in the figure.

        \textbf{Linearization by increasing $n$.}
        The distribution tends to linearize as $n$ increases, a trend observed in both native and non-native speakers. Since the corpus utilized in this experiment employs the same reading text for both groups, language fluency does not influence the linguistic content. Consequently, the linearization is anticipated to be driven by the textual information (particularly the semantic information hypothesized in \Sec{experiment-zipf}), rather than language fluency.
        
        \textbf{Deviation of non-native speakers.}
        When we examine the deviations of non-native speakers compared to native speakers, we notice deviations correlated with language fluency. Specifically, non-native speakers tend to use high-frequency symbols more frequently. These findings indicate that Zipf's law (and the power law) can be employed to discern differences between standard and non-standard speech with regards to language fluency. 
        Further exploration is required to ascertain the specific nature of these deviations. Intriguingly, speakers with high-level proficiency diverge more from native speakers. This counters the intuitive expectation that individuals with low-level fluency would deviate more significantly.

\section{Conclusion} \label{sec:conclusion}
    In this paper, we investigated whether speech symbols follow Zipf's law. Through our experiments, we determined that: 1) speech symbol $n$-grams corresponding to a word follow a power law rather than Zipf's law, and 2) non-textual deviations in non-standard speech can be identified through the power law of the rank-frequency distributions.
    
    Our research paves the way for textless analysis methods applicable to various audio data. The following are possible directions for future research.
    \begin{itemize} \leftskip -5mm
        \item \textbf{Language development.} As outlined in \Sec{introduction}, the Zipf's (or power) law can potentially aid in analyzing infants' vocabulary acquisition. Despite the existing challenges in robust automatic speech recognition of infants' voices~\cite{gebauer23_interspeech, gebauer23_interspeech}, our approach can encode voices into symbols without transcriptions. This enables statistical analyses utilizing voices exclusively.
        
        \item \textbf{Animal voices and non-speech audio.} Zipf's law has been demonstrated to apply to symbolic audio, including animal calls~\cite{arik21zipfslaw-animal-voice} and music scores~\cite{perotti20zipfslaw-music}. While current research often relies on human-invented symbols and annotations, our method could potentially extend to audio sources beyond human speech without using the human-invented symbols and annotations. General-purpose audio representation models~\cite{niizumi2021byol-a} could be employed for this purpose.
        
        \item \textbf{Emergent speech communication.} The analysis of emergent languages constitutes a significant area of research, with objectives including 1) fostering language-based machine-machine communication (i.e., communication between artificial intelligence)~\cite{serhii17nips_zipf, lazaridou2017multiagent}, and 2) evaluating the extent to which these communications mirror statistics of human-to-human communications~\cite{kottur-etal-2017-natural, harding-graesser-etal-2019-emergent}. Given that Zipf's law (and related principles, such as the Zipf's law of abbreviation~\cite{zipf2013psycho}) embodies the principle of least effort in human communication, it may offer a way to explore machine-machine communication~\cite{ueda21relationship}. Our method holds the potential to forge a way in examining speech-based machine-machine communication, potentially benefitting developments of exploring human spoken language emergence and human-machine communication.
    \end{itemize}

\printbibliography


\end{document}